\documentclass[letterpaper, 10 pt, conference]{ieeeconf}  

\IEEEoverridecommandlockouts                              

\usepackage{amsmath}
\usepackage{booktabs}
\usepackage{algorithm2e}
\usepackage{xcolor}
\usepackage{multicol}
\usepackage{subcaption}
\usepackage{multirow}

\usepackage{caption}
\usepackage{subcaption}
\usepackage{graphicx,subcaption}
\RestyleAlgo{ruled}
\SetKwComment{Comment}{/* }{ */}

\DeclareMathOperator*{\argmin}{arg\,min}

\title{\LARGE \bf
Dense Multi-Agent Navigation Using Voronoi Cells and Congestion Metric-based Replanning
}

\author{Senthil Hariharan Arul$^{1}$ and Dinesh Manocha$^{2}$
\thanks{*This work was not supported by any organization}
\thanks{$^{1}$Senthil Hariharan Arul is with Department of Electrical and computer Engineering, University of Maryland at College Park, Maryland, United States
        {\tt\small sarul1@umd.edu}}%
\thanks{$^{2}$Dinesh Manocha is with the Department of Computer Science, University of Maryland, Maryland, United States
        {\tt\small b.d.researcher@ieee.org}}%
}

\begin{document}

\maketitle
\thispagestyle{empty}
\pagestyle{empty}

\begin{abstract} 
We present a decentralized path-planning algorithm for navigating multiple differential-drive robots in dense environments. In contrast to prior decentralized methods, we propose a novel congestion metric-based replanning that couples local and global planning techniques to efficiently navigate in scenarios with multiple corridors. To handle dense scenes with narrow passages, our approach computes the initial path for each agent to its assigned goal using a lattice planner. Based on neighbors' information, each agent performs online replanning using a congestion metric that tends to reduce the collisions and improves the navigation performance. Furthermore, we use the Voronoi cells of each agent to plan the {\em{local}} motion as well as  a corridor selection strategy to limit the congestion in narrow passages. We evaluate the performance of our approach in complex warehouse-like scenes with hundreds of agents and demonstrate improved performance and efficiency over prior methods. In addition, our approach results in a higher success rate 
in terms of collision-free navigation to the goals.  
\end{abstract}


\section{Introduction}
Recent advances in AI and robotics have led to novel applications of these technologies in inventory management, delivery, and industrial automation. These applications can employ hundreds of robots~\cite{li2020lifelong} to transport between multiple source and destination locations. Recently, there has been considerable interest in using robots for warehouse automation.  Many organizations use large-scale fulfillment centers that organize the inventory using several movable pods. The pods get transported by a team of robots to different picking stations, where each item gets picked by a human operator.

It is important to develop efficient navigation algorithms that can be deployed in such dense and complex scenarios.
Multi-agent pathfinding (MAPF) is a key problem in  multi-robot systems and involves computing collision-free paths for the agents to their respective goal locations. 
Moreover, warehouse-like scenes are regarded as complex environments with narrow passages and dynamic obstacles corresponding to other robots and possibly humans. A key issue in terms of navigation is avoiding robot-robot and robot-human collisions. Most current solutions tend to use centralized planners to compute multi-agent paths. These centralized planners~\cite{liu2019task,honig2016multi} use global knowledge of all agent states and compute a path for each agent using a composite state-space that considers all the agents simultaneously. For optimal MAPF solvers, the state-space can grow exponentially with the number of agents~\cite{cbs}. This global knowledge can be used to generate optimal paths and can also provide some guarantees in terms of completeness but results in increased computation cost~\cite{jingjinyu_optimal}. Furthermore, these centralized methods may not work well when multiple robots get deployed in an unknown and unstructured environment, e.g., in scenarios with human co-workers or unknown dynamic obstacles. The state-of-the-art MAPF methods 
tend to generate piece-wise linear paths that involve ``stop and turn'' maneuvers and can be inefficient for non-holonomic agents~\cite{cohen2019optimal}. Moreover, the centralized computation can create a single point of failure~\cite{velagapudi2010decentralized}. 

In decoupled approaches agents plan their individual paths to improve scalability, while still using some global information similar to centralized methods. Decoupled approaches~\cite{ma2017lifelong,silver2005cooperative} use a block of memory containing the current paths of all agents, and the agents take turns replanning their paths. Sequentially planning agent paths affects completeness as one agent's path could block another agent's path to goal~\cite{silver2005cooperative}. 

\begin{figure}[t]
    \centering
    \includegraphics[trim={0.2cm 0.2cm 0.2cm 0.2cm}, clip, width=0.45\linewidth]{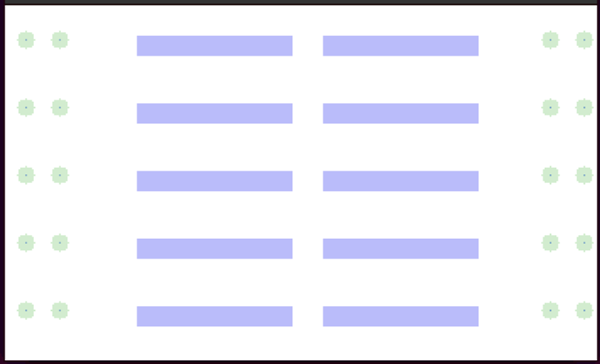}
    \hfill
    \includegraphics[trim={0.2cm 0.2cm 0.2cm 0.2cm}, clip, width=0.45\linewidth]{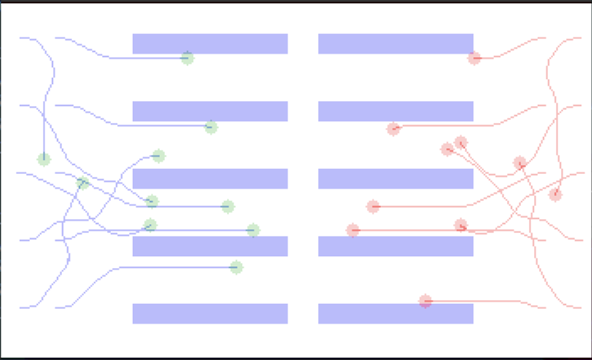}
    \caption{(Left image) Top-down view of a simple environment with static obstacles (blue polygon) and 20 agents (green disk). (Right image) Agents' paths for a scenario with 20 agents.}
    \label{fig:mesh1}
\end{figure}

In contrast, decentralized planners~\cite{VO,ORCA,zhou2017fast} use local information to navigate robots towards their goals. They improve scalability by trading path optimality and completeness for faster computation. In these planners, agents make independent decisions to avoid the collision and reach their goals using neighbors' state information. Since, decentralized methods\cite{ORCA,zhou2017fast} direct the robots in the goal direction, an agent can get into local minima in the presence of static obstacles. Hence, global planners are often used to generate initial paths to the goal, and decentralized methods follow the paths while avoiding the collisions locally~\cite{van2008interactive,LSwarm}.
Deadlocks are still possible as the agents could block each other's global plans, causing over-crowding or some agents may not be able to reach the goals.


{\bf{Main Results:}}
We present a novel algorithm for multi-agent navigation in dense, congested environments with hundreds of agents.  Our formulation is based on coupled global and local planning approach for collision-free navigation. Our method is decentralized, where each agent replans its path independently based on local observation of its environment. We use lattice planning as the basis for generating smooth paths for differential drive agents by accounting for static obstacles. Our main contributions include:
\begin{itemize}
    \item We present a novel formulation for decentralized multi-agent navigation that combines lattice-based replanning and Voronoi diagram computations.
    \item We propose a {\em{congestion metric}} to select between the alternate paths to the goal from the agent's current location. A suitable path is chosen that is considered {\em{safe}} based on the congestion metric. Our metric considers path distance, potential collisions, and agent crowding while performing conflict resolution and reducing deadlocks. 
    \item To alleviate crowding or congestion, we propose an approach to handle narrow passages by allowing each agent to reserve passageways using a shared database.
\end{itemize}

We evaluate our method in simulated environments with narrow passageways and observe significant improvement in performance over prior decentralized strategies. Our evaluation metrics include path length, success rate in reaching the goal, deadlocks, and collisions observed.  In addition, our proposed method scales with the number of agents owing to their decentralized nature. 
\section{Related Work}
Multi-agent navigation is a well-studied topic in robotics involving the collision-free navigation of multiple agents towards their respective goals. Multiple methods have been proposed in this respect, and the problem is NP-hard in terms of computing optimal paths~\cite{ma2017overview}. Please refer to~\cite{Felner2017SearchBasedOS} for a detailed survey. In general, multi-agent navigation methods are classified into centralized or decentralized planners. 

\subsection{Centralized Planners}
Centralized planners consider a composite state-space including all the agents in the workspace~\cite{tang2018complete}. The composite state-space provides global information about the workspace to compute collision-free paths to the goal. Centralized planners produce smooth, near-optimal paths that are guaranteed to be collision-free. On the other hand, they are not scalable to a large number of agents due to their large state space and centralized computations~\cite{solovey2016finding,goldenberg2014enhanced}. Centralized methods like push and swap~\cite{Luna2011PushAS} are complete and computationally efficient but may not generate optimal paths. Further, some planners assign priority to agents to break conflicts and improve scalability, as in Cooperative A*~\cite{silver2005cooperative} or Priority Based Search (PBS)~\cite{ma2018searching}. Yu and LaValle~\cite{jingjinyu_optimal} present an algorithm for generating optimal makespan solutions for multi-robot planning in challenging problems using a mapping to a multiflow network.

\subsection{Decentralized Planners}
In decentralized planners, each agent independently computes a path that locally avoids collisions. Velocity obstacle(VO)~\cite{VO} computes a set of collision-free velocities based on the position and velocity of the agent and its neighbors. The VO formulation was extended to consider reciprocity in RVO~\cite{rvo} and formulated as a linear programming problem in ORCA~\cite{ORCA}. The formulation is extended to a variety of agent dynamics, including double integrator agents~\cite{AVO}, differentiable drive~\cite{orcadd}, and non-holonomic agents~\cite{nhorca}. Buffered Voronoi Cell (BVC)~\cite{zhou2017fast} is an efficient decentralized method that can compute collision-free trajectories for single integrator agents. In contrast to VO-based formulations, BVC only requires the agents' position information to compute collision-free paths. RVO and BVC is used for multi-agent navigation in~\cite{vrvo}. 

\subsection{Multi-Agent Path Finding}
Hybrid planners combine the advantages of centralized planners and decoupled computation. Sharon et al.~\cite{cbs} present Conflict Based Search (CBS), a two-level planner for optimal multi-agent pathfinding. CBS constructs a conflict tree (CT) in which each node contains a set of constraints. CBS checks for conflicts among its paths and resolves them by splitting CT nodes into child nodes. ECBS~\cite{ecbs} enhances the scalability of CBS to a large number of agents by using a bounded sub-optimal planner. 
CBS-based planners are extensively used in  warehouse planning. These methods consider a grid structure environment and compute a piecewise linear path. Since warehouse robots are primarily non-holonomic, they can result in inefficient maneuvers such as turning in place. Kinematic constraints are considered~\cite{honig2016multi,Ma_Honig_Kumar_Ayanian_Koenig_2019} to compute smooth and efficient paths. Guo et al.~\cite{jingjinyu_splitting} propose spatial and temporal splitting schemes to increase scalability of methods like ECBS on challenging maps, with negligible effects on path optimality.

Above MAPF methods plan using a centralized server, and the scalability is limited due to the centralized computation. Decoupled planners~\cite{ma2017lifelong,silver2005cooperative} use token passing where a block of memory is communicated between the agents. This block of memory consists of the current planned path of each agent, and the agent currently holding the token is allowed to replan and update the memory.

Geraerts and Overmars propose a corridor map~\cite{corridormap} method for generating smooth, short paths for multiple agents in real-time. He et al.~\cite{interpolationbridge} propose computing bridges with geometric properties to enable the computation of collision-free paths in environments with narrow passages. Park et al.~\cite{jongjinpark} propose a cost based evaluation of agent trajectories similar to our proposed method. 

Recently, game-theoretic planners have been gaining traction in the multi-agent navigation community. Spica et al.~\cite{spica2018realtime} present a planner for two-player autonomous drone racing that showcases competitiveness amongst agents. Mylvaganam et al.~\cite{mylvaganam2017differential} present a differential game approach to multi-agent collision avoidance.  

In contrast to the above method, we present a decentralized conflict cost-based replanning method that uses a lattice structure and BVC to plan a smooth, collision-free path. Though our method plans on a lattice map, which is similar to grid maps used in prior MAPF methods, we resolve possible collisions locally using our congestion metric-based replanning. Moreover, the corridor selection helps with reducing the congestion and deadlocks.  
\section{Problem Formulation and Background}
In this section, we provide an overview of our approach. Table~\ref{tab:booktabs} summarizes the symbols and notations used in the paper. 
\begin{table}
\centering
\caption{Symbols and Notations}
\begin{tabular}{lrr}
\toprule
Notation  & Definition \\
\midrule
$A_i$ & Refers to the $i^{th}$ agent\\
 $t_i$ & Task $i$ which is an element of task set T\\
 $\mathbf{p}_{i}$ & Current 2-D position of $A_i$ $[p_{i,x}, p_{i,y}]$\\
 $\mathbf{s}_i$ & Initial configuration of $A_i$\\
 $\mathbf{g}_i$ & 2-D goal position of task $t_i$\\
 $\mathbf{v}_i$ & 2-D velocity of $A_i$ $[v_{i,x}, v_{i,y}]$\\
${R}_i$ & Radius of $A_i$'s enclosing circle \\
$\mathcal{N}_i$ & Neighbors of $A_i$\\
$\mathcal {V}_i, \bar{\mathcal {V}}_i$ & Voronoi and Buffered Voronoi cell for $A_i$ \\ 
$\partial\bar{\mathcal {V}}_i$ & Boundary of the set $\bar{\mathcal {V}}_i$ \\
$Env$ & Environment model for lattice planner\\
$\pi({\bf{s}},{\bf{g}})$ & Path between configurations {\bf{s}} and {\bf{g}}\\
\bottomrule
\end{tabular}
\vspace*{-0.13in}
\label{tab:booktabs}
\end{table}

\begin{figure}[t]
    \centering
    \includegraphics[trim={11cm 0cm 0cm 4cm}, clip, width=0.8\linewidth]{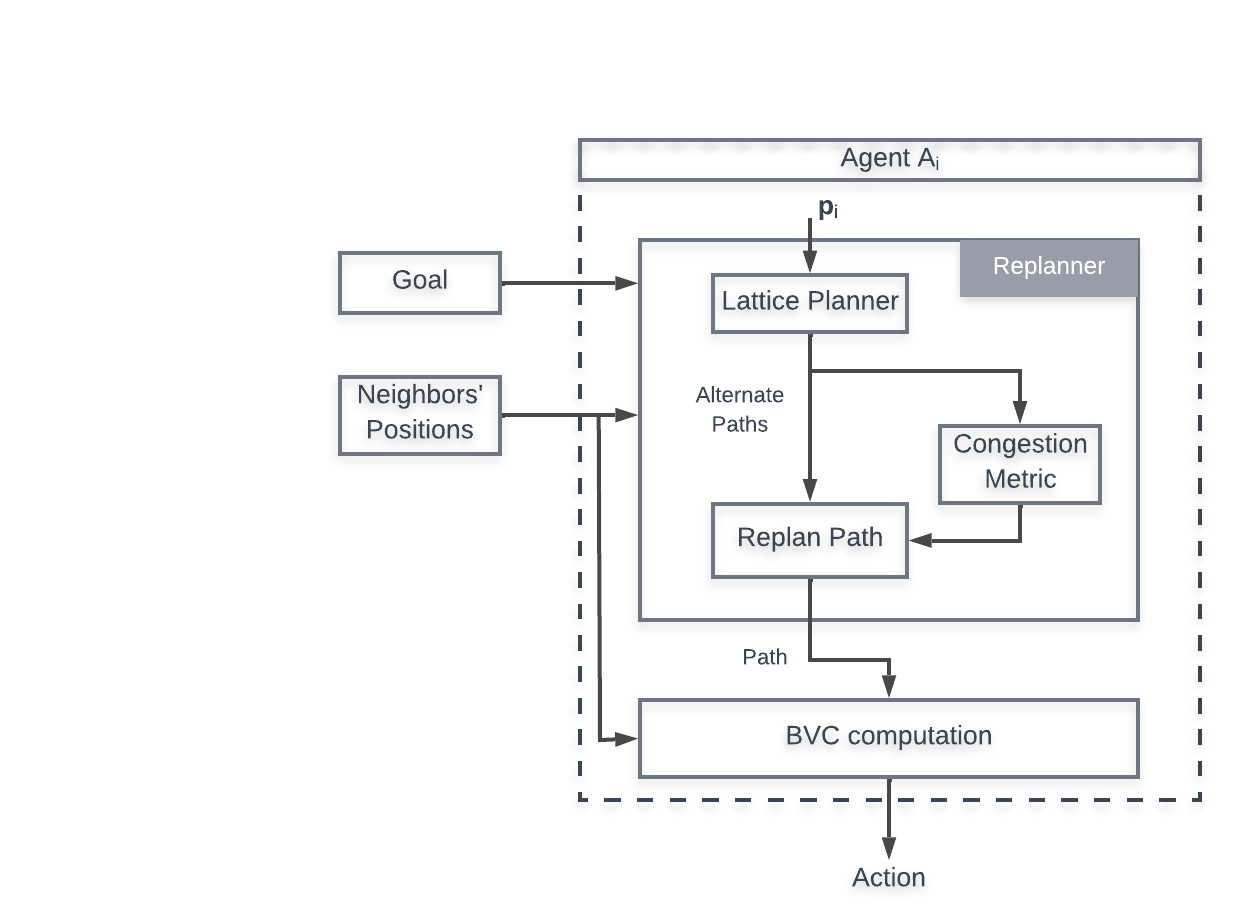}
    \caption{Our multi-agent replanning method:. Each agent received its goal position and neighbors' positions as an input. A set of alternate paths to the goal are generated considering the possible trajectories from the current position. We compute congestion metrics for each path and each agent chooses   path that minimizes the congestion metric. Each agent follows the chosen path such that its local trajectories are limited within its BVC.
}
    \label{fig:framework}
\end{figure}

\subsection{Problem Statement}
We consider the problem of generating multi-agent paths for collision-free navigation. We consider a workspace $\mathcal{W} \subset \mathbf{R}^2$ with $\mathbf{N}$ homogeneous agents. In this work, we limit our approach to disk-shaped agents with differential drive dynamics. We consider a pair of agents $A_i$ and $A_j$, for $i,j \in \{1, 2, ..., N\}$ to be in collision if their bounding disks overlap, that is,
$$\Vert \mathbf{p}_i - \mathbf{p}_j \Vert_{2} \le R_i + R_j.$$
We consider that an external task allocator assigns the goals (2-D coordinates) to the agents. In many dense scenarios, a decentralized planner can result in deadlocks as agents can block each other's paths or the planner can get stuck in a local minimum. Moreover, we use the success rate of a multi-agent system in term of reaching the goal configuration as a suitable metric to evaluate the planner's performance. 


\subsection{Assumptions}
In our approach, we consider disk-shaped, homogeneous agents. We assume each agent $A_i$ knows its neighbors' positions, where neighbors are agents located within a certain radius of $A_i$. This information can be obtained using perception or communication modules or using a Vicon setup. The agents' operating environment is known, and the environment's lattice discretization is precomputed. 

Since we resolve collision using local information, agents moving in narrow passageways could result in congestion. A representative scenario would be a narrow passageway of single-agent width. Agents entering the passageway from opposite direction could result in a deadlock. Thus, in addition to our planning method, we propose using a reservation table of the narrow passageways in the environment. A set of narrow corridors is identified and updated in a central database accessible by all agents. During execution, these passageway can be reserved for a single direction motion (Section~\ref{subsection:corrselect}). 

\subsection{Lattice Planning}
Lattice-based motion planners sample the feasible state space of the robot in a regular fashion to construct a graph structure~\cite{Pivtoraiko-2005-9311}. The graph has feasible states as its nodes and feasible motions primitives as the edges. In this paper, we consider non-holonomic constraints, and the graph $G=(V,E)$ embedded in Euclidean space presents the feasible, static, obstacle-free poses with feasible motion as the edges. We precompute a lattice graph that accounts for the static obstacles in the environment, and the agents use {{AD*}~\cite{ad*}}/ARA*~\cite{ara} algorithm to compute paths to their respective goals. 

\subsection{Buffered Voronoi Cell (BVC)}\label{BVC}
BVC~\cite{zhou2017fast} is a contracted Voronoi region generated by retracting the Voronoi edges by a distance equal to the agent's bounding radius. Disk-shaped agents remain collision-free provided their centers reside inside or on the boundary of BVC. Given $\mathbf{N}$ agents on a 2-D plane, the buffered Voronoi cell corresponding to an agent $A_i$ is given as,
\begin{equation}
    \bar{\mathcal {V}}_i=
    \bigg\lbrace {\bf p}\in \mathbf {R}^2| \bigg ({\bf p}-\frac{{\bf p}_i+{\bf p}_j}{2}\bigg)^\text{T}{\bf p}_{ij} + R_\text{i}\Vert {\bf p}_{ij}\Vert \leq 0, \forall j\ne i\bigg\rbrace. 
\end{equation} 
Here, ${\bf{p}}_{ij} = {\bf{p}}_i - {\bf{p}}_j$. The method plans a collision-free path for the agent $A_i$ that is constrained to lie within $\bar{\mathcal{V}}_i$.


\section{Multi-Agent Replanning}
\begin{figure}[t]
    \centering

    \includegraphics[trim={0.5cm 0.0cm 0cm 0cm}, clip, width=0.95\linewidth]{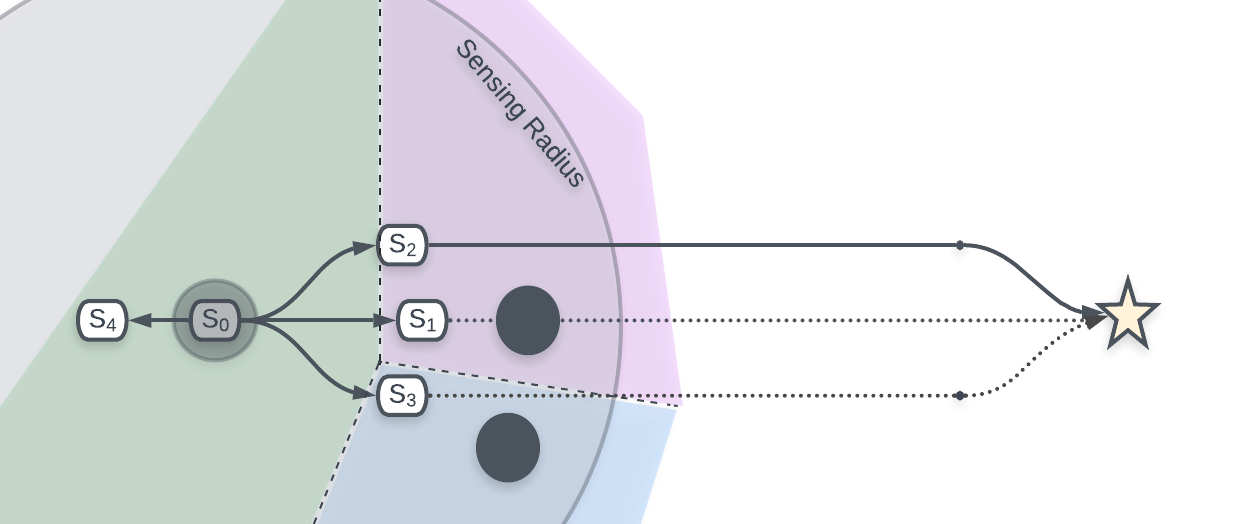}
    \caption{We highlight a navigation scenario with 3 agents. The ego agent, represented by a grey disk, is tasked with navigating towards the goal, represented by a star. The ego agent has two neighbors in its sensing region, both represented by black disks. The ego agent at state ${\bf{s}}_0$ has 4 successor configurations given by ${\bf{s}}_1, {\bf{s}}_2, {\bf{s}}_3$ and ${\bf{s}}_4$. The potential paths to the goal through the successor nodes are computed and are shown in the figure. We observe that the paths through ${\bf{s}}_1$ and ${\bf{s}}_3$ leads to collisions, and hence their congestion metrics are larger than for ${\bf{s}}_2$. Hence, the ego vehicle chooses the path through ${\bf{s}}_2$. The agent's local motion is limited by its BVC, and its Voronoi cell is highlighted as a green region in the figure. As the agent moves, its Voronoi cell changes and the agent remains collision-free, provided it remains within its BVC. The Voronoi cells for the obstacles are represented by blue and purple regions.}
    \label{fig:conflict}
\end{figure}

In this section, we summarize our planning algorithm, the congestion cost computation, and our replanning approach. Our overall approach is summarized in Algorithms~\ref{alg:two} and~\ref{alg:three} and a high-level overview is illustrated in Figure~\ref{fig:framework}. A three-agent scenario is illustrated in Figure~\ref{fig:conflict}.

{\subsection{Global Plan Generation}}
We consider a system with $N$ homogeneous agents with individual goals (for each agent) assigned through an external task allocator. We represent a task using a 2-tuple $(t_i, \bf{g}_i)$. Here, $t_i$ represents the task id, and $\bf{g}_i$ represents the goal configuration. Our planning algorithm utilizes a lattice structure as its basis to compute the initial global plan and the subsequent re-plan. The lattice graph for the environment is precomputed.

Each agent $\mathcal{A}_i$ individually computes a smooth path between its source configuration $\mathbf{s}_i$ and its goal configuration $\bf{g}_i$ using the environment layout. The lattice graph structure is queried to identify a suitable path using the {\color{black}{AD*}~\cite{ad*}} planner. Since the graph edges are precomputed motion primitives, the computed path is feasible for the agent. Moreover, the path is collision-free with the static obstacles in the environment. The computed initial global path is represented by $\pi_{i}({\bf{s}}_i,{\bf{g}}_i)$, which represents a sequence of 3-tuples given by $(\bf{p}_{i,x}, \bf{p}_{i,y}, \theta_i)$ between the agent's source ($\bf{s}_i$) and its goal ($\bf{g}_i$) configuration.

\subsection{Collision Avoidance Using BVC and Congestion Metric Replanning}
The computed global path $\pi_{i}({\bf{s}}_i,{\bf{g}}_i)$ is not guaranteed to be collision-free as it is oblivious to inter-agent collisions or collisions with dynamic obstacles. As the agent traverses this path, its local environment changes, and collisions may be imminent due to neighboring agents. Considering the agent's current position ${\bf{p}}_i$, the subsequent portion of the path to goal is given by $\pi_i({\bf{p}}_i,{\bf{g}}_i)$. 
We employ a cost function to intelligently choose between multiple possible paths between ${\bf{p}}_i$ and ${\bf{g}}_i$ to avoid collisions and congestion. We formulate this as an optimization problem to choose a desired global path, which minimizes the metric. 

The main goal of the congestion metric is to perform the task of collision avoidance while moving the agent towards its goal. In this regard, the metric must account for the possibility of collision in the path, the path cost, and density/crowding in the path. A good path must have a low probability of collision, a small path cost, and should not move too close to its neighbors. Our metric accounts for these criteria using three costs, as mentioned below.

Let us assume that agent $A_i$ is at a location $\mathbf{p}_i$. The agent at a lattice node can choose $n$ different motion primitives ($prim_k \forall k \in {1,2,...n}$), which are the edges of the lattice graph. The agent following the primitives will reach subsequent lattice nodes represented by ${\bf{p}}_{prim_k,i}$. Considering each ${\bf{p}}_{prim_k,i}$ as the starting configuration, we compute $n$ paths to the goal. The $n$ paths are represented by $\pi_{prim_k,i} \forall k \in {1,2,...n}$. 
\begin{equation*}
    \pi_{prim_k,i}({\bf{p}}_{i}, {\bf{g}}_{i}) \gets \pi_i({\bf{p}}_{i}, {\bf{p}}_{prim_k,i}) + \pi_i({\bf{p}}_{prim_k,i}, {\bf{g}}_i)
\end{equation*}
The set of $n$ paths is represented by,
$$
\Pi_i = \bigcup_{k=1}^{n}  \pi_{prim_k,i} ({\bf{p}}_i,{\bf{g}}_i).
$$

We choose a suitable path from the set $\Pi_i$ such that the agent avoids collision, reduces congestion, and approaches its goal. We rank the $n$ paths based on their congestion metric, which is given by
$$
C(\mathcal{\pi}_{prim_k,i}) = k_c * C_{collision} + k_g * C_{goal} + k_n * C_{neighbor},
$$
where $k_c$, $k_g$, and $k_n$ are positive weights for the individual costs.

\begin{itemize}
    \item Collision Cost ($C_{collision}$): This captures the cost of colliding with an obstacle. Based on the neighbors' positions, and velocity (${\bf{p}}_j, {\bf{v}}_j \text{  } \forall j \in \mathcal{N}$), we identify collisions in the current path using $\pi_{i}$. For each neighbor lying in the current path, we add a positive value to the cost.
    \item Goal Cost ($C_{goal}$): This computes the increase in the path length due to replanning and promotes choosing the shorter path. 
    \item Crowding Cost ($C_{neighbors}$): To reduce agent crowding, we use a neighbor cost term. The neighbor cost is computed as,
    \[
    C_{neighbors} = 
    \begin{cases}
    \frac{1}{t*||{\bf{p}}_i - {\bf{p}}_j||},& \text{if } ||{\bf{p}}_i - {\bf{p}}_j|| \leq 5R_i\\
    0,              & \text{otherwise}
    \end{cases}
    \]
    Here, $t$ is the approximate time instant at which the agent would reach ${\bf{p}}_i$ from its current position using $\pi_i$. 
\end{itemize}

Since the agent has a finite set of motion primitives at each node, it is easy to compute the cost for each possible path. We choose a global path $\pi_i({\bf{p}},{\bf{g}})$ that minimizes the congestion.
$$
\pi_i({\bf{p}}_{i}, {\bf{g}}_{i}) \gets  \argmin_{\pi \in \Pi_i}{C(\mathcal{\pi})}
$$
    


Using the congestion metric, the agent chooses a suitable path to maneuver towards the goal. 

To reduce the possibility of collision, we use BVC to limit the progress of the agent along the computed paths. The agent constructs a buffered Voronoi cell (BVC) based on its position (${\bf{p}}_i$) and its neighbors' positions. The BVC indicates the collision-free configurations for the agent, and if all agents continue to move within their BVCs, the agents will remain collision-free. Each agent follows its global path such that it lies within its BVC.

$$
{\bf{p}}_i \in \mathcal{\bar{V}}_i \implies || {\bf{p}}_i - {\bf{p}}_j|| > R_i + R_j, \text{   } \forall j \in \mathcal{N}_i
$$
At each time step, the agent moves ahead along its path if the next position is with in the BVC. The BVC is update at the next time step based on the updated local information. The process is continued until the robot reaches its goal.


\begin{algorithm}[t]
\small
\caption{Proposed Multi-Agent Navigation Method}\label{alg:two}
\bf{Input:} \tt{Robot State, Env}\;
\bf{Output:} \tt{Action}\;

$\pi_i \gets \tt{LatticePlanner(Env, {\bf{s}}_i, {\bf{g}}_i)}$\;
\While{$\bf{s}_i \neq \bf{g}_i$}{
    ${\bf{p}}_{new,i} = \tt{nextPosition}(\pi_i)$\;
    $\mathcal{\bar{V}}_i \gets \tt{computeBVC(Robot Positions)}$\;
    \If{${\bf{p}}_{new,i} \in \mathcal{\bar{V}}_i$}{
    ${\bf{p}}_i = {\bf{p}}_{new,i}$\;
    }
    $\pi_i({\bf{p}}_{i}, {\bf{g}}_{i})$ = \tt{conflictResolution()}\;
    
}
\end{algorithm}

\begin{algorithm}[t]
\small
\caption{\tt{conflictResolution()}}\label{alg:three}
$\Pi$ = $\pi_i({\bf{p}}_{i}, {\bf{g}}_{i})$\;
\For{$x \in successor({\bf{p}}_i)$}{ 
\Comment*[r]{successor is a node reachable from current node following a single motion primitive}
$\pi_{temp} \gets \tt{LatticePlanner(Env, \bf{x}, \bf{g}_i)}$\;
$\Pi \gets \Pi \cup \pi_{temp}$\;
}
$\pi_i({\bf{p}}_{i}, {\bf{g}}_{i}) \gets \argmin_{\pi \in \Pi}{C(\pi)} $\;
return $\pi_i({\bf{p}}_{i}, {\bf{g}}_{i})$\;
\end{algorithm}


\section{Deadlock Reduction and Handling Narrow Passages}\label{subsection:corrselect}


\begin{figure}[t]
     \centering
     \begin{subfigure}[t]{0.45\linewidth}
         \includegraphics[trim={0.1cm 1.1cm 0.1cm 0.1cm}, clip, width=\textwidth]{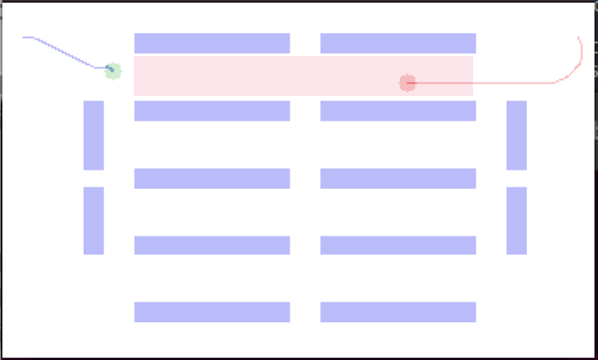}
         \subcaption{(a) Corridor reserved by the red agent. The corridor is highlighted in red.}
         \label{fig:cs1}
     \end{subfigure}
     \hfill
     \begin{subfigure}[t]{0.45\linewidth}
         \includegraphics[trim={0.2cm 1.1cm 0.1cm 0.2cm}, clip, width=\textwidth]{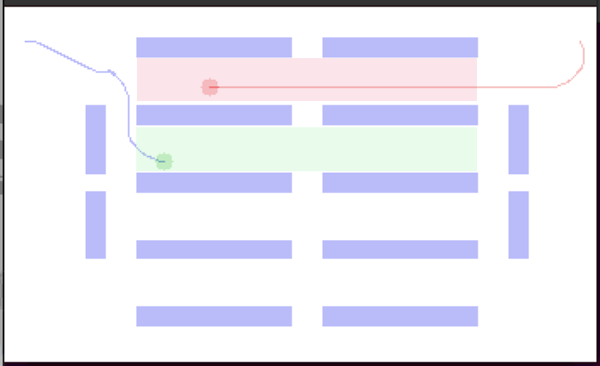}
         \subcaption{(b) New corridor (green) chosen by the blue agent to reduce congestion.}
         \label{fig:cs2}
     \end{subfigure}
     \hfill
     \vfill
    \begin{subfigure}[t]{0.45\linewidth}
         \includegraphics[trim={0.1cm 1.1cm 0.1cm 0.1cm}, clip, width=\textwidth]{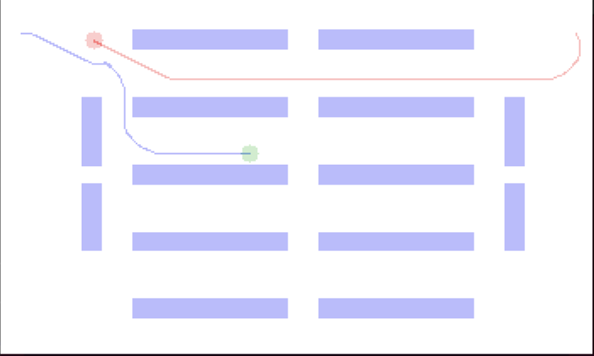}
         \caption{(c) Final agent paths.}
         \label{fig:cs3}
     \end{subfigure}
        \caption{Two agent scenarios representing our corridor selection algorithm. The initial paths for both agents pass through the top corridor. Since the agents arrive from opposite directions, we reserve the corridor dynamically in a particular direction.}
        \label{fig:three graphs}
\end{figure}

Environments consisting of narrow passageways can result in deadlocks and longer paths, as agents may try to traverse the passage in opposite directions. To reduce such inefficiencies, we restrict any narrow passage to single direction traffic at any given time. As a preprocess, we create a database of all narrow passageways in the environment. During runtime, all agents can access this shared database to reserve passageways or to check a passageway's reservation status. 

An agent whose path traverses a passageway can reserve it for unidirectional travel on reaching a certain radius around the entry. Other agents can traverse the reserved corridor provided its traversal direction aligns with the reserved traversal direction for the passageway. That is, agents cannot enter the passageway in the opposite direction.

Each passageway in the warehouse layout is identified and represented by a unique identifier. Let us say the layout consists of $n$ passageways represented by $\mathcal{P}_i$  $\forall i \in \{1,2,...,n\}$. For each passageway $\mathcal{P}_i$ we define a parameter\\ $\tt{reserved_{\mathcal{P}_i} = (status, direction, start time, end time)}$. A shared database holds this block of data and is accessible by all agents.\\
Based on the reservation status, we have three cases.
\begin{itemize}
    \item \textbf{Case 1: Passageway unreserved}\\
    Agents can select an unreserved passageway to move in a particular direction. Based on the agent's estimated arrival time at the passageway's entry, the passageway is reserved for a computed time duration. Since the agent knows the distance to the passageway's entry from its global path and the average velocity, we have a lower bound estimate of the earliest arrival time. Similarly, the reservation end time is computed using a duration estimate based on the passageway length and the average agent speed.
    \item \textbf{Case 2: Passageway reserved in same direction}\\
    Other agents intending to use the passageway in the reserved direction can update the reservation end time based on an estimate of time duration required to traverse the passageway.
    \item \textbf{Case 3: Passageway reserved in opposite direction}\\
    In this case, we create temporary obstacles at the entry of the passageway and replan a global path. Thus, the agent will compute a path to its goal using a different passageway.
\end{itemize}

Following the above algorithm, we have a global path that is followed using our collision avoidance rule. The above algorithm is repeated until the agent reaches the goal.

\section{Results}

\subsection{Experimental Setup}
Our algorithm was implemented on an Intel Xeon octacore processor at 3.6 GHz with 32 GB memory and GeForce GTX 1080 GPU. We use the SBPL~\footnote{{http://sbpl.net/Home}} package with AD* planner for the lattice planning. The motion primitives are constructed considering a unicycle dynamics for the agent. 


\subsection{Benchmarks}
\begin{figure}[t]
\begin{subfigure}[t]{0.48\linewidth}
    \includegraphics[trim={0.1cm 0.1cm 0.1cm 0.1cm},clip, width=0.9\linewidth]{images/topDown}
    \subcaption{(a) Environment 1}
\end{subfigure}
\hfill
\begin{subfigure}[t]{0.48\linewidth}
    \includegraphics[trim={0.1cm 0.1cm 0.1cm 0.1cm},clip,width=0.9\linewidth]{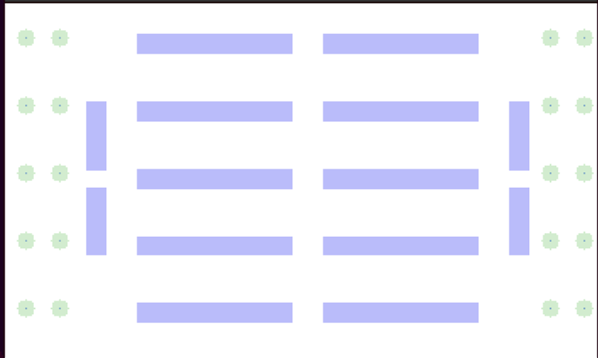}
    \subcaption{(b) Environment 2}
\end{subfigure}
\caption{Scenarios and Environment}
\label{fig:env}
\end{figure}

\begin{figure}[t]
    \centering
        \includegraphics[width=0.8\linewidth]{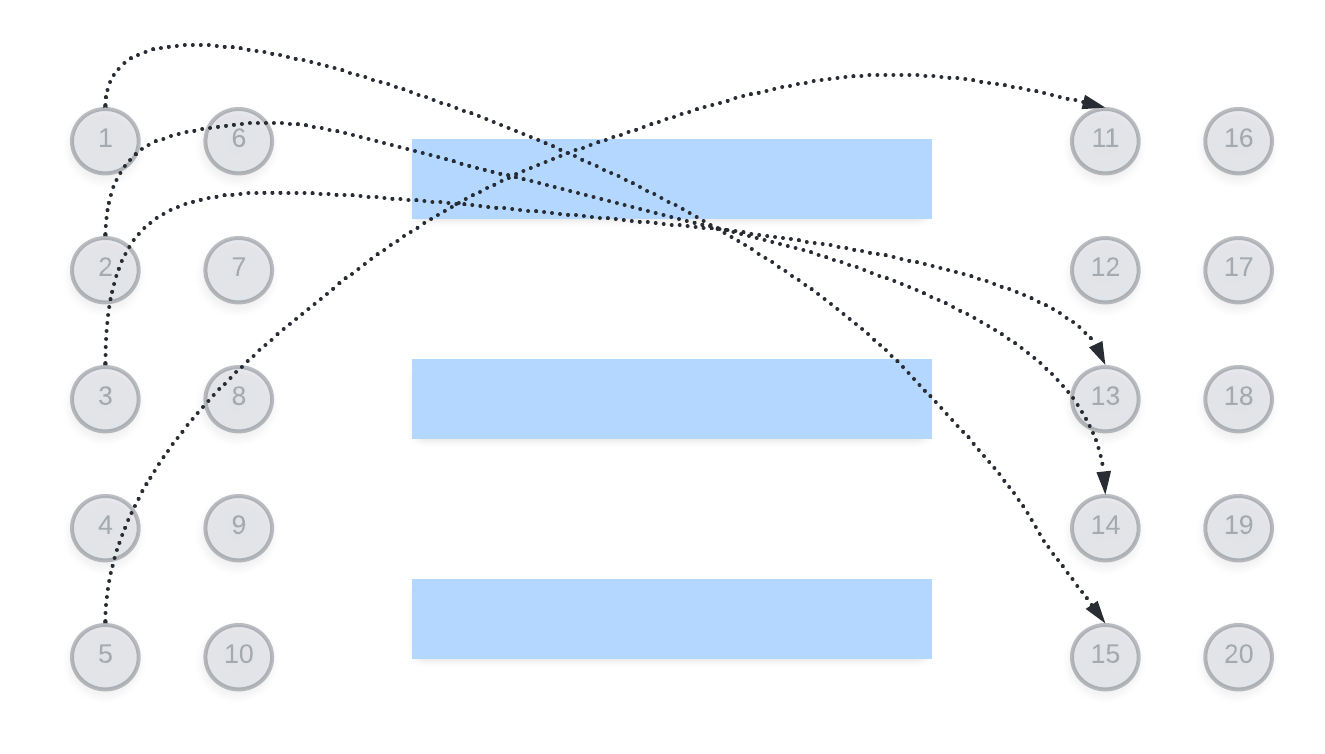}
    \hfill
    \caption{Evaluation Scenarios: The task involves the agents moving in towards their opposite direction such that the agents in the top rows move towards the bottom, as illustrated.}
    \label{fig:evalScenario}
\end{figure}

We evaluate our algorithm in simulated environments with multiple passageways. The environments are illustrated in figure~\ref{fig:env}. The agents are tasked with navigating from their configuration to a final configuartion as illustrated in figure~\ref{fig:evalScenario}.

\subsection{Evaluations}

\begin{figure*}[t]
    \begin{subfigure}[t]{0.24\linewidth}
         \includegraphics[trim={0.07cm 0.05cm 0.05cm 0.07cm}, clip, width=\textwidth]{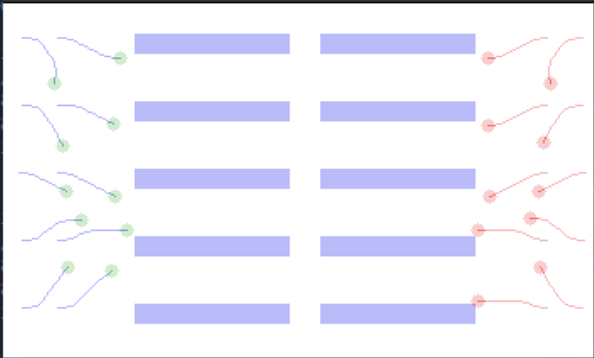}
         \subcaption{(a)}
         \label{fig:cs1}
     \end{subfigure}
     \hfill
     \begin{subfigure}[t]{0.24\linewidth}
         \includegraphics[trim={0.07cm 0.12cm 0.05cm 0.07cm}, clip, width=\textwidth]{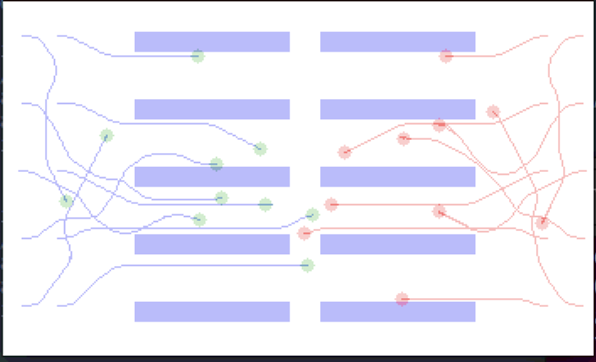}
         \subcaption{(b)}
         \label{fig:cs1}
     \end{subfigure}
     \hfill
     \begin{subfigure}[t]{0.24\linewidth}
         \includegraphics[trim={0.05cm 0.05cm 0.05cm 0.07cm}, clip, width=\textwidth]{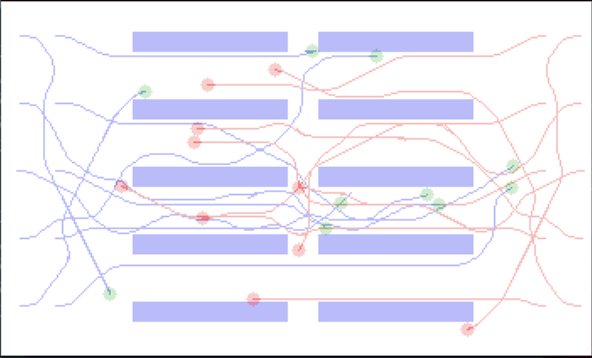}
         \subcaption{(c)}
         \label{fig:cs1}
     \end{subfigure}
     \hfill
     \begin{subfigure}[t]{0.24\linewidth}
         \includegraphics[trim={0.05cm 0.05cm 0.05cm 0.1cm}, clip, width=\textwidth]{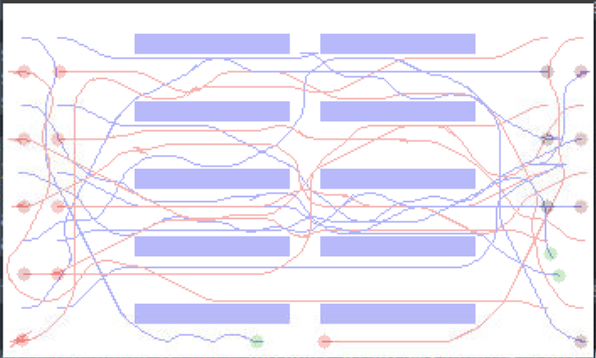}
         \subcaption{(d)}
         \label{fig:cs1}
     \end{subfigure}
\caption{Agent paths for 20 agents in a simulated warehouse environment.}
\label{fig:traj}
\end{figure*}

\begin{table*}[t]
  \centering
  \begin{tabular}{|c||c|c|c|c|c|c|c|}
    \hline
    \multirow{2}{*}{Environment}
    & & \multicolumn{2}{c|}{AD* Planner} & \multicolumn{4}{c|}{Proposed Method}\\
    \cline{3-8}
    & Agents & Agents Collided & Path Length & Agents Deadlocked & Agents Collided & Success Rate & Path Length \\
    \hline
    \multirow{2}{*}{\shortstack{Environment 1\\ (Vertical Agent Spacing 4m)}}
    &  5 &  4  & 37.77 & 0 & 0 & 1 & 42.50\\
    & 10 &  9  & 37.20 & 0 & 0 & 1 & 41.13\\
    & 20 &  20 & 36.95 & 0 & 0 & 1 & 41.96\\
    & 30 &  30 & 38.13 & 0 & 0 & 1 & 43.99 \\
    \hline
    \multirow{2}{*}{\shortstack{Environment 1\\ (Vertical Agent Spacing 2m)}}
    & 20 & 19 & 36.99 & 0 & 0 & 1 & 41.03\\
    & 40 & 40 &  37.69 & 0 & 2 & 0.95 & 44.67\\
    & 60 & 60 & 37.36 & 0 & 4 & 0.93 & 45.36\\
    \hline
    \multirow{2}{*}{\shortstack{Environment 2\\ (Vertical Agent Spacing 4m)}}
    &  5 & 0  & 41.82 & 0 & 0 & 1 &  42.71\\
    & 10 & 6  & 40.04  & 0 & 0 & 1 &  41.97 \\
    & 20 & 20 & 40.19 & 0 & 0 & 1 &  44.54 \\
    \hline
    \multirow{2}{*}{\shortstack{Environment 2\\ (Vertical Agent Spacing 2m)}}
    & 20 & 19 & 40.07 & 0 & 0 & 1 & 43.13\\
    & 40 & 40 & 40.16 & 0 & 0 & 1 & 46.57\\
    \hline
    \end{tabular}
  \caption{Evaluation Scenarios}
  \label{tab:1}
\end{table*}

\subsubsection{Trajectories}
In Figure~\ref{fig:traj}, representative trajectories for 20 agents are highlighted in the evaluation environment. In contrast to navigation algorithms which produce piece-wise linear paths, our planner produces trajectories that are smooth owing to the lattice planner used.

\subsubsection{Success rate, Collisions, and Deadlock}
We defined success rate as the fraction of the agents reaching their goal while avoiding collision. The success rate ($SR$) is defined as,
$$
SR = 1 - DF - CF.
$$
Here, $DF$ is the fraction of agents stuck in a deadlock, and $CF$ is the fraction of agents that collided. Episode that have both collision and deadlocks are accounted in $CF$.
Table~\ref{tab:1} summaries the success rate of our proposed method in dense environments.

\subsubsection{Path Length}
In Table~\ref{tab:1} we summaries the average path length of the agents in the evaulation scenario. We tabulate the path length while the agent uses a AD* star planner alone with our collision avoidance. This provides a comparision of the incrwease in path length from the indivudual agents path,

\section{Conclusion, Limitations, and Future Work}

We introduced a decentralized multi-agent path planner to navigate robots to their respective goals in dense environments. We evaluated our algorithm against the decentralized method in a set of comprehensive scenarios in terms of path length, success rate, etc. Our algorithm shows improved performance in a variety of scenarios. 

Our method has a few limitations. We currently consider a homogeneous system, and all the agents use the same navigation strategy. Moreover, we require the positions and velocity of the neighboring robots to be known. The corridor selection requires a stable communication channel between the robot and the central server, and performance could decline due to communication lag. Our future work would be to generalize the algorithm for any dynamics, consider human obstacles, and test the algorithm on physical robotic systems.





\bibliographystyle{IEEEtran}
\bibliography{IEEEabrv,IEEEbib}

\end{document}